# Surprise! Using Physiological Stress for Allostatic Regulation Under the Active Inference Framework [Pre-Print]

Imran Khan and Robert Lowe
Digitalization, Interaction, Cognition, and Emotion (DICE) Lab
Department of Applied IT
University of Gothenburg
Gothenburg, Sweden
imran.khan | robert.lowe @ ait.gu.se

*Abstract*— Allostasis proposes that long-term viability of a living system is achieved through anticipatory adjustments of its physiology and behaviour: emphasising physiological and affective stress as an adaptive state of adaptation that minimizes long-term prediction errors. More recently, the active inference framework (AIF) has also sought to explain action and long-term adaptation through the minimization of future errors (free energy), through the learning of statistical contingencies of the world, offering a formalism for allostatic regulation. We suggest that framing prediction errors through the lens of biological hormonal dynamics proposed by allostasis offers a way to integrate these two models together in a biologically-plausible manner. In this paper, we describe our initial work in developing a model that grounds prediction errors (surprisal) into the secretion of a physiological stress hormone (cortisol) acting as an adaptive, allostatic mediator on a homeostatically-controlled physiology. We evaluate this using a computational model in simulations using an active inference agent endowed with an artificial physiology, regulated through homeostatic and allostatic control in a stochastic environment. Our results find that allostatic functions of cortisol (stress), secreted as a function of prediction errors, provide adaptive advantages to the agent's long-term physiological regulation. We argue that the coupling of information-theoretic prediction errors to low-level, biological hormonal dynamics of stress can provide a computationally efficient model to long-term regulation for embodied intelligent systems.

I. INTRODUCTION

A. *Background*

In both biological and artificial systems, mechanisms of adaptation are critical to long-term stability and viability in dynamic, unpredictable environments. Long-term stability of living systems has historically been explained through homeostasis [1], [2], which proposes that (life-essential) physiological parameters come equipped with a fixed range of critical limits (attractor states) that must be maintained within to ensure survival. Control of these internal parameters is achieved by correcting deviations through negative feedback loops, propagating error signals (deviations from set points) through the system to drive corrective action that restores a parameter back to its set point (or within its set critical range). Under this definition, however, homeostatic control is an inflexible and metabolically inefficient means towards long-term stability [3]. Not only is reacting after the event potentially catastrophic to the organism but fixed homeostatic set points are bootstrapped to specific environmental contexts, leaving little room to adapt to changing, unpredictable, and stress-inducing environments.

The concept of allostasis [3], [4] proposes instead that stability is achieved not (strictly) through local, reactive error-correction, but through feedforward, anticipatory action through prediction of future bodily needs. Rather than just reacting to deviations from prior attractor states (homeostatic set points), allostasis proposes that both the internal parameters (including homeostatic set points) as well as behaviours of an organism are adjusted ahead of potential future deviations, by trying to predict the future state of dynamic (internal and external) environments. These predictions are generated from prior experiences, applied to current (internal and external) context: calibrating the internal homeostatic system to environmental demands, proposing a more energy-efficient approach to long-term regulation than reactive homeostasis [5].

Deviations between prior (allostatic) predictions and sensory inputs result in *prediction errors*, leading to correction either through (error-correcting) homeostatic or (error-preventing) allostatic adjustments. These prediction errors drive allostatic adjustments: periods of acute, adaptive, physiological (and psychological) "stress" [3], mobilising and adapting numerous physiological and behavioural responses to re-establish homeostatic balance, e.g. a vasodilation response in anticipation of the need to run away from a predator. Thus, physiological affective states of stress are not strictly maladaptive, but play a critical adaptive role in maintaining homeostatic balance in unpredictable environments. These adaptations are mediated through stress-related hormones: glucocorticoids, including cortisol, acting as *coordinators of environmental responsiveness* [6] (p. 2). Indeed, canonical theories of allostasis [2], and more recent work [7], [8], [9] propose hormonal signalling (particularly via cortisol) as a key mediator of anticipatory physiological and neural adjustments Physiological and affective stress, therefore, is necessary for long-term adaptability.

Though allostasis offers a plausible, biologically grounded concept for long-term regulation, ambiguity in its usage and a lack of a formal model have likely hindered its wider-spread adoption. More recently, the active inference framework (AIF) [10] offers a way to recast allostasis in formal terms. Like allostasis, AIF seeks to explain an agent's interoceptive control, cognition, and even affect (e.g. [11], [12]) through minimising prediction errors: formally defined in AIF as an information-theoretic quantity known as *free energy*. Here, agents are considered "prediction machines" [13], equipped with a set of prior preferences for sensations (e.g. internal set points) and beliefs about the world (a generative model). Agents seek to predict the causes of the sensory signals that they receive (from the internal and external milieu), seeking to generate sensations (through perception and action) that can best fit prior beliefs and future preferences (thus, minimising future prediction errors).

AIF offers a robust formalism from which to approach adaptive and allostatic regulation [14], [15], [16] as well as affective regulation [5], [11], [17], [18] of (biological) systems. However, where AIF proposes adaptation and error minimisation to (partially) occur through updating probabilistic distributions of a generative model, allostasis emphasises hormonally driven adjustments and modulation of underlying physiological systems. These low-level, hormonally driven changes underpin physiological and psychological affect (e.g. stress [7]) in biological systems, and can offer metabolically and computationally-efficient [19] ways of internally representing environmental dynamics across multiple time scales in embodied agents. This permits adaptivity and long-term viability amidst changing environmental demands.

To this end, we argue that coupling information-theoretic prediction errors (from AIF) with affective regulation of physiology (via hormonal dynamics) from allostasis can be expressed through a bio-inspired artificial system. In this paper, we propose a model that integrates anticipatory, allostatic regulation of physiology under the active inference framework, using a physiological stress-related hormone (cortisol), tied into the dynamics of information-theoretic prediction errors (or *free* energy) and uncertainty[1] over future actions (variance in *expected free energy*). Cortisol provides two allostatic functions in this model: firstly, by adjusting the prior target value (homeostatic set point) of a life-essential physiological variable, calibrating physiology to perceived environmental demands. Secondly, by modifying the rate of the agent's learning of the statistical contingencies of the world (i.e. learning of its generative model).

We evaluate the adaptive properties of this model using a computational model of an active inference agent endowed with an artificial physiology interacting with a dynamic stochastic environment. With this work, we aim to contribute an additional perspective in the recent work that seeks to integrate allostasis, AIF, and affect [14], [15], [16] through the use of physiological (affective) hormonal dynamics.

In Section I.B. below, we contextualise our model and aim to provide unfamiliar readers with a (high-level) introduction to the active inference framework (however, given our limited space, we refer readers to [10], [20] for more detailed descriptions). We describe the specific implementation of our active inference agent in our simulations and describe our experiments and results in Section II and Section III respectively. We finish with our discussion of these results, including limitations and areas of future work in Section IV.

*B. Active Inference Framework*

The active inference framework (AIF) is a recent theory in computational neuroscience that seeks to unify (and explain) perception, action, and learning of (biological) organisms [7], [16]. It extends existing Bayesian approaches to perceptual inference (e.g. predictive coding [22]) by accounting for action, and predicting future consequences of taking actions. Under AIF, agents are equipped with a set of prior preferences for sensory data and beliefs about the world. Perception, action, and learning are all in service of fulfilling their prior preferences, specifically by minimizing the discrepancies (also known as *surprisal*, formally measured as *free energy*) between these prior predictions or expectations, and the data that is being observed from the environment.

AIF starts with the assumption that all organisms are equipped with a set of prior preferences for sensory inputs (which can be psychological, or hard-wired as phenotypic/evolutionary preferences, e.g. homeostatic set points for body temperature). For instance, biological agents typically prefer internal signals indicating that they are fed, to maintain body temperature within a certain range, and so on. AIF further assumes that all agents represent their world through an internal (generative) model consisting of prior beliefs (probabilistic distributions) over different causes of sensory input. Here, causes of observations are hidden (latent): the true state of the world cannot be directly observed by the agent. Therefore, by sampling observations (sensory signals) and using prior probabilistic distributions over the potential causes of those observations, the generative model is inverted (through variational inference) to *infer* (i.e. find a posterior distribution over) the most (probabilistically) likely cause (state) of each observation.

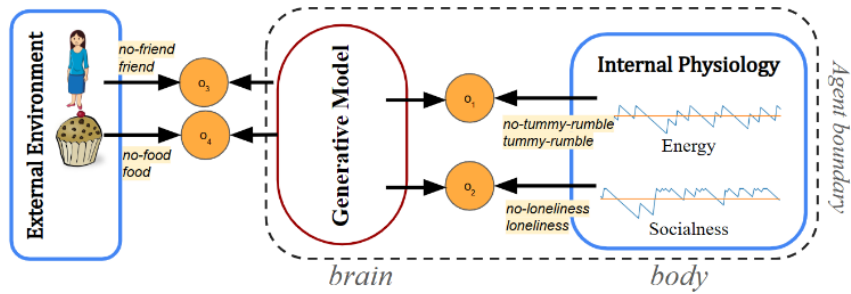

Fig. 1. Overview of the agent – comprised of the generative model and an internal physiology – and the external environment. The "agent" boundary denotes our definition of the agent in this paper. Note: the terms "brain" and "body" are used for illustrative purposes.

Like allostasis, *prediction errors* are discrepancies between an agent's prior beliefs or expectations (e.g. an expectation for homeostatic balance) and the information it senses from the environment (e.g. interoceptive signals related to dyshomeostasis). Agents seek to minimize prediction errors/free energy through a number of different (but concurrent ways): (a) through sampling of additional observations from the environment, (b) updating its beliefs (her generative model) about the world to attempt to better (approximately) match them to (the true state of) the environment, and (c) acting on the environment to change their observations, i.e. to generate sensory data that match prior beliefs or preferences. Actions are selected based on the minimization of prediction errors in the future (*expected* free energy) under a given action or policy. In sum, agents attempt both to update their internal model (beliefs) of the world they interact with, and to pursue actions which they believe (based on their current understanding of the world) will result in smaller prediction errors (*surprisal*) in the future.

## II. AGENT SIMULATION

### A. Simulation Environment

We build and test our proposed models in a computational environment. Our simulation environment consists of two components: (a) a simulated agent, endowed with an artificial physiology (a "body"), regulated via homeostatic (and allostatic) control, along with a probabilistic (generative) model (its "brain") that performs inference over the causes of its sensory data and selects the "optimal" action to maintain homeostasis; and (b) an external environment that (randomly) generates two types of resources—which we call *food* and *friend*—which the agent can perceive and can take action on to satisfy its internal variables. The goal of the agent is to simply "stay alive" by maintaining its life-critical internal variables above their target values, through the interoceptive or exteroceptive regulation of its physiology. The agent must infer the (hidden) state (its **motivation state**) that it believes causes the interoceptive (and exteroceptive) signals that it observes. Note that, in these simulated models, these terms are illustrative, since our agent is a disembodied, computational agent. Fig. 1 illustrates our agent and environment set up.

### B. External Environment

The external environment is the world beyond the boundaries of the agent (Fig. 1): a non-spatial environment representing a "physical" world from which the agent can receive observations and interact with. The external world consists of two resources (*food* and *friend*) whose presence is represented as discrete variables (1 if they are observed, else 0). In our simulations, resources are randomly generated (with 20% probability) at each time step.

---

1 Whilst AIF has its own definition of "uncertainty", we use it in this paper as a measure of "indecisiveness" over future actions. We discuss this more in Section II.F

*C. Agent Physiology and Action*

The agent's physiology consists of two internal physiological parameters $v$, which we simply call *Energy* ($v_1$) and *Socialness* ($v_2$). These represent two competing internal needs, aligning to the classic, ethology-inspired "Two-Resource Problem" [23] which has also been employed in [19], [24]. These values take the range 0-1. Each of these two internal needs have a corresponding attractor state or homeostatic set point $d$ (initially set to 0.7). Like biological organisms, these variables incur small deficits over time, moving away from ideal (attractor) states and towards states that threaten viability and life (i.e. running out of *Energy*):

$$v_{n,t} = v_{n,t-1} - \gamma_n \quad (1)$$

where $v_{n,t}$ is the value of internal variable **v** at time **t** and $\gamma_n$ is the rate of loss (here, set at **0.03** for both variables). The agent aims to maintain its viability by keeping each of these internal needs satisfied (at or above the homeostatic set point, i.e $v_n \geq d_v$) through either error-correcting action (homeostasis), or anticipatory/predictive adjustments (allostasis). Note that when the agent's *Energy* drops to 0, it will die.

Non-satisfaction (and satisfaction) of these physiological parameters is represented as categorical variables, generating one of two interoceptive signals for each variable **v**. We refer to these signals as *"tummy rumble"* and *"loneliness"* when each of these variables drop below their respective target values/homeostatic set points (i.e. $v_n < d_v$). We describe this in more details in Section II.D. To correct these deficits, the agent selects from one of three available actions (*eat, play, explore*), where action selection is driven by the generative model, based on the minimisation of future expected prediction errors (expected free energy or surprisal). We describe these behaviours in more detail in Section II.E.

*D. Generative Model*

As we described in Section I.B., an agent's perception and action are determined by the *generative model*: a set of beliefs (probabilistic distributions) an agent has about the (true) state of the world (i.e. *"the causal structure of the environment and how that structure generates observations"* [12] p.2). Our generative model is formulated as a partially-observable Markov decision process (POMDP), formally defined as **p(o,s,u)**, consisting of a set of discrete/categorical observations (**o**), causes of those observations (latent states) (**s**) and possible actions (**u**) that the agent can take. It also encapsulates probabilistic distributions (beliefs) over the possible causes of observations, **p(o|s)**, beliefs about how states change over time, **p(s_t|s_{t-1}|u_{t-1})**, and a prior distribution (preferences) over observations **p(o|C)**. In line with existing work [20], these are represented as arrays in our model which we refer to as A, B, and C array, respectively. Details can be found in Table 1.

TABLE I.  Model parameters of the generative model with formal definitions and descriptions

| Model Parameter | Formal definition | Description |
|---|---|---|
| A Matrix | $p(o\|s)$ | Sensory likelihood mapping: the probability distributions that each hidden state **s** caused observation **o**, i.e. *"what is the likelihood that each motivation state **s** caused the interoceptive and exteroceptive data **o** that I have observed?"* |
| B Matrix | $p(s_t\|s_{t-1}, u_{t-1})$ | Transition distribution describing the probability of transitioning to a state **s** at time **t** given state **s** and action **u** at **t-1**, i.e. *"what is the likelihood that my motivation state is **s** given (my beliefs) about the state I was just in and the action that I just took?"* |
| C Vector | $p(o\|C)$ | Set of preferences for each observation, i.e. *"what sensory (interoceptive/exteroceptive) signals do I prefer to observe?"* |

**Observations:** Our agent receives four observation modalities **o**: two **interoceptive** signals (called "*tummy*" and "*lonely*") related to each of its internal needs ("*Energy*" and "*Socialness*"), and two **exteroceptive** signals (related to the presence of *food* or a *friend*). Interoceptive signals are generated from the agent's physiology (Section II. C), whilst exteroceptive signals are generated by the external world (Section II.B).

**Hidden State:** Our model contains a single hidden state, **s**: an agent's **motivational state**. Since the agent cannot directly know its (latent) motivational state, it must infer it based on its beliefs about the causes of sensory (interoceptive and exteroceptive) signals (i.e. through its generative model). The agent can be in one of three motivational states: *hungry, playful, or satisfied.* In the initial model, the sensory likelihood matrix (A array) is initialized such that being *hungry* has a high probabilistic cause of observing *food* and/or *tummy-rumble*; being *playful* is a likely cause of observing *loneliness* and/or *friend*: being *satisfied* is the likely cause of observing *no-tummy-rumble* and *no-loneliness*. This is akin to the cue-deficit model of motivation as put forth by [25].

**Action Selection:** The generative model selects an action **u** which can best (probabilistically) minimize future prediction errors i.e. which can best match the agents' prior expectation to maintain homeostasis. This is formally achieved by scoring each action based on how well it has minimized prediction errors/free energy up to this point and combining it with how much it is expected to reduce prediction errors (surprisal) in the future. Actions that are expected to minimize future prediction errors (i.e. that are most likely to generate preferred sensory signals) are selected. Here, the agents' preference for sensory signals corresponds to biological survival: it prefers to see interoceptive signals of being *no-tummy-rumble,* and, to a lesser extent, *no-loneliness*. Action selection is then in the service of the agent predicting which actions can best lead to these interoceptive signals.

**Learning:** Finally, the agent can update its generative model (i.e. beliefs about the world) over time. Specifically, our agent updates prior parameters of the state transition probabilities (B Matrix) based on posterior beliefs at each time step (see footnote 2 for full mathematical details). Here, the initial learning rate is set to 0.05 and modulated by cortisol (Section II. F).

## E. Agent Behaviours

When an action **u** is selected by the generative model (Section II.D), the agent attempts to enact the action onto the external world. The possible actions are: *eat* a *food* resource, *play* with a *friend* (both of which are consummatory behaviours) or *explore* the world (an appetitive behaviour). Since the external world generates resources randomly (20% probability), successful execution of consummatory behaviours is only possible when a resource is available (i.e. during the same time step that the agent selects to perform that action). The exception is *explore*, which, when selected, guarantees the observation of both resources in the next time step (i.e. both *food* and *friend* will be observed at *t+1*). Each of the two consummatory behaviours, *eat* and *play*, satisfy the internal variables *Energy* and *Social*, by a value of **+0.4**: *explore* does not contribute to changes in internal variables.

TABLE II. OBSERVATION MODALITIES, STATES, AND ACTIONS IN THE GENERATIVE MODEL. THE INITIAL PROBABILISTIC DISTRIBUTIONS OVER ALL OF THESE CAN BE SEEN IN OUR SUPPLEMENTARY MATERIAL.

| | |
|---|---|
| **Observation Modalities** | $o^{Tummy} \in \{$ no-tummy-rumble, tummy-rumble $\}$<br>$o^{Lonely} \in \{$ no-loneliness, loneliness $\}$<br>$o^{food} \in \{$ no-food, food $\}$<br>$o^{friend} \in \{$ no-friend, friend $\}$ |
| **Hidden States** | $s^{motivation} \in \{$ hungry, playful, satisfied $\}$ |
| **Actions** | $u^{behaviour} \in \{$ eat, play, explore $\}$ |

## F. Simulating Cortisol Secretion and Allostatic Functions

The secretion of cortisol is a function of two components within the active inference framework. Firstly, it is a function of the changes in prediction error (*surprisal*): that is, it is a measure of the difference between predicted observations and the actual sensory inputs at a given time step. Secondly, this secretion is modulated by the level of uncertainty over the best action to take. We define uncertainty here as the difference between the best and worst actions (in terms of minimizing expected free energy), i.e. indecisiveness of how to best minimize future prediction errors. To summarize, cortisol is secreted when i) an agent's current observations deviate from its expectations, and ii) according to the level of 'indecisiveness' about what the best course of action is to correct these deviations. This is formalized accordingly:

$$CT_t = CT_{t-1} + (-ln(p(o_t)) - -ln(p(o_{t-1}))) + (1 - (G_{max} - G_{min})) \quad (2)$$

Where $-ln(p(o_t)) - -ln(p(o_{t-1}))$ is the difference in surprise between the previous and current time step, and $(1 - (G_{max} - G_{min}))$ is the (normalized) difference between the maximum and minimum posterior over actions. Here, the less "confident" an agent is that one action is the most "optimal" (in terms of minimizing expected free energy), the more cortisol will be released. Cortisol has two allostatic functions in the model. First, it adjusts the internal homeostatic set point (i.e. the internal value associated with the prior preference **p(o|C)**) associated with the life-critical variable *Energy*:

$$d_{energy, t} = d_{energy, t-1} / (1 + CT_t - CT_{t-1}) \quad (3)$$

Where $d_{energy}$ is the set point of *Energy* at time t, and CT are the cortisol levels as calculated in (2). On a slower time-scale, cortisol modulates the learning rate for updating state transition probabilities (Table 1, B matrix) in the generative model. As this learning is considered to be analogous to Hebbian learning in AIF [26], this is an abstraction of cortisol's effect on synaptic plasticity and learning [27]. Trivially[2], this can be represented as:

$$P(B) = B + (\lambda * (1 - CT)) \qquad (4)$$

Where *P(B)* is the posterior B matrix, and $\lambda$ is the predefined learning parameter (0.05).

Fig. 2. Overview of the four models tested in our simulations. GM = Generative Model. IP = Internal Physiology. B = B Matrix (Table 1). Pink arrows denote additions to each model compared to previous ones.

### III. EXPERIMENT SET UP

*A. Experimental Conditions*

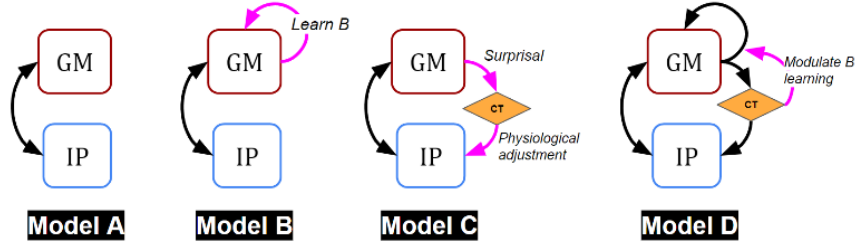

We test four different models related to different allostatic effects of cortisol which can be seen in Fig. 2. In **model A**, we use a model of simple homeostatic regulation of an internal physiology through active inference (i.e. the generative model infers motivational states and optimal action selection). In **model B**, we extend the first model by adding a simple learning component (as described in Section II.D) of state transition dynamics (the B Matrix), with a predefined learning rate (determined by pre-experimental calibration). These two simulations serve as our control models. In **model C**, we extend the homeostatic model A and incorporate an allostatic effect of cortisol (Section II.E), through adjustments of the homeostatic set point for *Energy* (3). Finally, in **model D**, we extend the latter model by also incorporating cortisol's modulatory function to the learning rate associated with the state transition matrix in the agent's generative model (4).

In all conditions, the initial generative model is pre-defined and provided to the agent. We summarise the setup of this generative model in Section II and provide this initial distribution in our supplementary material.

Simulations were written and performed using Python 3.10 primarily using the *pymdp* library [28]. Simulations were run for 300 iterations; 10 simulation runs per model. Given the stochasticity of the environment, we removed simulations where each resource was not generated at least twice in the first 50 time steps. We evaluate each model primarily on its **viability** (defined as simulation time where *Energy* > 0) and the deviation of both physiological variables from their set points, as a measure of how well the agent is managing its internal physiology (**physiological comfort**, where higher values are better). We also report on mean **cortisol** levels throughout the length of time that the agent remained viable. We provide qualitative assessments of the dynamics of (posterior) state beliefs and action selection in each model. We also provide visualisations of the rate of change in surprisal over the simulations, to visualise the dynamics of the agents internal "affective valence", akin to [11], [29].

### IV. RESULTS

Table 3 provides a summary of the mean results of each model across all 10 simulations. We provide raw data of our results in our supplementary material. In the subsequent figures, we present representative outputs of the internal variables of each model (Fig. 3). We also provide visualisations of state beliefs and action selection (Fig. 4).

---

[2] Formally speaking, this is achieved through modulation of the learning rate of the Dirchelet parameters over the transition model B. A full mathematical breakdown of this process can be found at [20] p.31.

TABLE III. SUMMARY OF RESULTS (WITH SEM) FOR ALL MODELS TESTED. DETAILS OF EACH MODEL CAN BE FOUND IN SECTION III.

| Metric | Model A | Model B | Model C | Model D |
|---|---|---|---|---|
| **Viability (/100%)** | 41±7% | 68±7% | 89±4% | 96±4% |
| **Action Distribution % Eat/Play/Explore** | 13/87/0% | 32/33/35% | 15/85/0% | 35/32/33% |
| **Median Comfort** | 49±16% | 75±5% | 96±1% | 81±6% |
| **Cortisol (Mean)** | .27 ± .04 | .35 ± .13 | .30 ± .03 | .49 ± .12 |

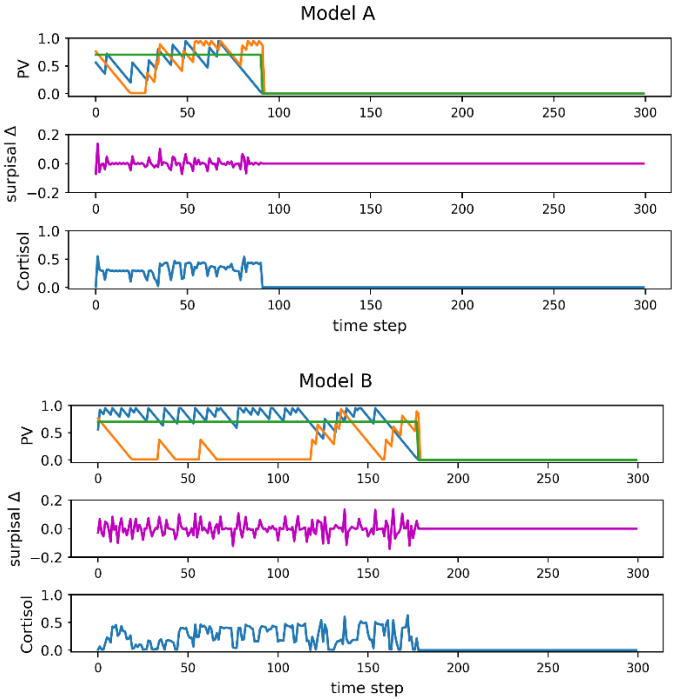

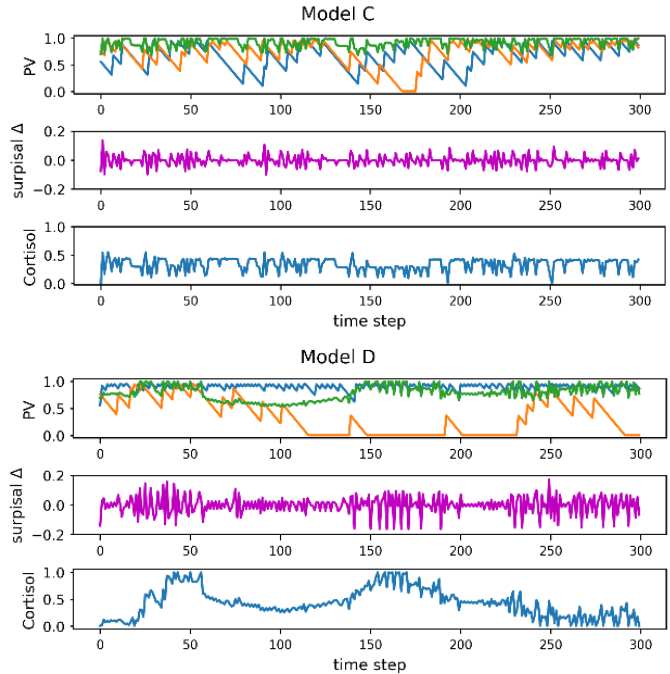

Fig. 3. Representative outputs of the internal variables (PV) (blue = *Energy*. Orange = *Socialness*. Green = Homeostatic set point for *Energy*), surpisal delta/affective valence, and cortisol dynamics for each model.

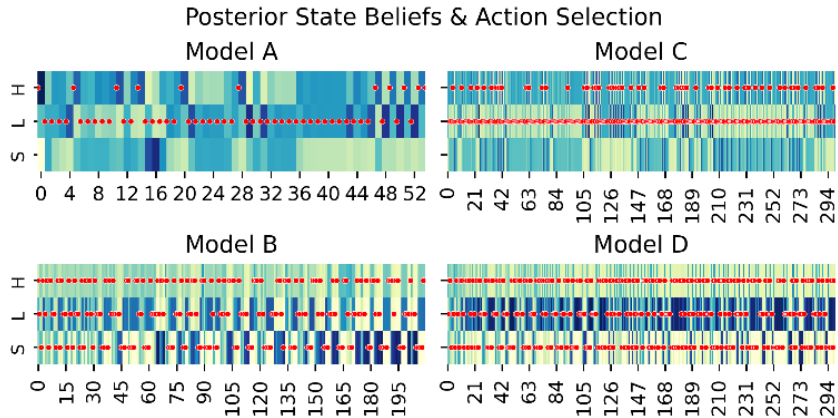

Fig. 4. Representative outputs of the agent's beliefs about its motivational state (H = Hungry, L = Lonely, S = Satisfied), graded from light yellow (low probability values) to dark blue (high probability values). Red dots denote the selection of the behaviour related to that motivation (eat for hungry, play for lonely, explore for satisfied).

### A. Model A: Homeostatic Active Inference Agent

In **model A**, the agent maintained its viability for 123 time steps on average (41±7%) of simulation run time, with a reported mean physiological comfort of 49±16% (69±10% for *Energy* specifically). From all the different models tested, this was the poorest performing model in terms of physiological regulation metrics. In terms of action selection: despite the agent being endowed with an initial generative model that left room for *exploration*, (see Supplementary Material) the *explore* action was not selected by the generative model across any of the simulations, even at times that the model held strong beliefs that the motivation state was *satisfied* (e.g. time steps 12-17, and 20-25 in Fig. 4). Instead, the agent persisted to choose consummatory policies, choosing the *play* policy (87% of the time), and only pursuing the *eat* policy (13%) at time steps where it had very strong beliefs about its *hunger* state (Fig. 4, model A). Though cortisol had no effect in this model, it reported a mean value of .27±.01 across all simulation runs.

*B. Simulation B: Homeostatic AIF Agent with Learning*

In **model B,** we extended the model in model A by adding a simple learning parameter to the agent's generative model (i.e. updating its beliefs about its state transition model). Here, the agent maintained its viability for 68±7% of the 300 time steps across the 10 simulations, reporting a median physiological comfort of 75±5%. The life-essential *Energy* variable was held at approximately 111±20% of its target set point (i.e. 11% above its target value of 0.7). Action selection was more varied than in model A: consummatory *eat* and *play* policies were selected 32% and 33% of the time, respectively. Unlike model A, this model also selected to *explore* the environment (35% of total action selection). Like model A, cortisol had no effect in this model, but reported a value of mean value of .35±.13. We discuss these points further in the Discussion section. We note that the addition of the learning component allowed the agent to become more confident in its beliefs of its motivational state over the course of the simulation (Fig. 4, Model B).

*C. Model C: Allostatic AIF Agent (No Learning)*

**Model C** incorporated allostatic adjustment of the target set point for life-critical *Energy* via cortisol (3). The agent maintained its viability for 267 time steps (89±4%) on average in these simulations, maintaining a median physiological comfort of 96±.1% of both variables. *Energy* was maintained to within 82±6% of its (allostatically-adjusted) target value. Over all simulations, this prior target value was adjusted from its initial value of .7 to .81±.2 (a 15% adjustment). In this model, mean cortisol levels were consistent, with a value of between .27--.33. Much like in model A, which also included no learning, the agent exclusively pursued the *play* (85% of action selection) and *eat* (15% of action selection) actions: never choosing to *explore*, despite times when it had strong beliefs that it was in the *satisfied* state (e.g. Fig. 4, model C, time steps 210-273).

*D. Model D: Allostatic AIF Agent with Learning*

The final **model D** extended model C by adjusting the learning rate of the transition model (B matrix). Here, the agent maintained its viability at 96±4% of simulation time (approximately 288/300 time steps). Physiological comfort for both internal variables were 81±6%: for *Energy* specifically, comfort was at 108±7%: like model B, *Energy* was mostly sustained slightly above its target value, even when this value was dynamically adjusted via allostatic mechanisms. Here, the set point for *Energy* increased by ~8.5%: from .7 to .76±.02 over the simulations. However, as seen in Fig. 2, the second (non life-essential) variable *Socialness* was poorly maintained in these simulations. Mean cortisol values in this model were .41±.12: the largest values across any of the models. Like model B, action selection was roughly evenly distributed: 35% for *eat*, 32% for *play*, and 33% for *explore*. In this model, the agent typically became confident about its beliefs of internal states early in the simulation run (Fig. 3, Model D).

## V. Discussion

Overall, the results of our simulation found that mechanisms of a stress-related hormone (cortisol), acting as an allostatic mediator on agent physiology as well as on learning of the generative model (models C and D respectively), provided significant advantages to the long-term, homeostatic regulation of an active inference agent: both in terms of the time it remains viable, as well as the appropriate management of competing internal drives. These effects were seen in active inference models both without (models A and C) and with learning (models B and D). We also find that allostatic adjustment of physiology (model C) outperformed active inference agents endowed with learning (model B). Similarly, we find physiological adjustment to be a less "costly" (in terms of cortisol secretion) approach to long-term regulation than with adapted learning (model D). We discuss some of our results in further detail below.

*A. Improved Regulation via Allostatic Adjustment of Homeostatic Set Points*

We find that allostatic adjustments of a prior homeostatic set point, as a function of prediction error or "stress" (via cortisol), allowed the agent to better regulate their homeostatic system. Our control model (model A) emerged to behave as a quintessential homeostatic model, i.e. "only eat when absolutely required". Likely given the stochasticity of the environment and resource availability, this strategy resulted in a viability time of only 41%. When cortisol provided an adjustment to the internal target value for *Energy* (model C), agent viability more than doubled to 89% of simulation time, with over twice as high comfort (45% vs. 96%). These beneficial effects of physiological adjustment were also observed in models that included learning (models B vs D).

The dynamics of the allostatic adjustments across time scales can be seen in Fig. 2 (model D). On a short time scale (50 time steps), early rises in cortisol levels downregulated learning, whilst simultaneously adjusting the prior target value of *Energy* to reflect the uncertainty of the stochastic environment. Despite the simple model of cortisol used here, this move allowed the agent to successfully adapt its physiology to the initial perceived demands of the world: it did not see what it "phenotypically" expected to see and so establishes a new set point. This new set point drives increased *eat* actions, keeping *Energy* elevated. This then "makes room" (e.g. frees up energetic resources) for the learning of the generative model (B matrix) occurring on a slower time scale. As a better model of the world is being learned (time steps 50-150), cortisol is inhibited, allowing *Energy's* set point to shift to lower values. This allowed the agent to anticipate the potential demands of the environment early in the simulation: protecting its "survival" first and foremost by keeping *Energy* elevated, while simultaneously learning about the world to promote its long-term viability.

*B. The Performance of Cortisol as a Signal of Environmental Dynamics*

We expected the performance of the homeostatic model (model A) to be improved when learning (model B) was introduced to the generative model: learning about the world is, naturally, advantageous. However, we found that the adjustment of *Energy's* set point tested in model C resulted in better viability and regulation of homeostasis than in simulations where the agent was able to update the statistical contingencies in the generative model (model B). Here, cortisol, acting as an allostatic mediator, served as a better functional signal for capturing the environmental dynamics, allowing the agent to "calibrate" to the uncertainty of the environment, and provided the agent with a better ability to maintain its homeostatic balance in the long term. This finding adds to [19] who also found hormonal dynamics to serve as an adaptive "biomarker" of environmental dynamics in embodied artificial agents. Here, we show that an active inference agent endowed with these (low-level) adaptive, affective hormonal dynamics is capable of outperforming one with learning capabilities.

*C. Contextualising Viability by its Cortisol "Cost"*

Despite the improved viability performance in models where learning was implemented (in particular, model D), these models reported the largest mean cortisol values (model B, .35; model D, .41) of all the models tested. Though cortisol has (simple) allostatic effects in our model, elevated and prolonged levels of stress and cortisol are maladaptive in biological entities. In reality, allostatic adaptation is costly, incurring wear-and-tear ("allostatic load") [30], underpinning chronic pathophysiological states. Therefore, the performance of these models should be contextualized with respect to the cortisol "cost" needed to achieve it. To this end, although the performance of models C and D were similar in terms of physiological regulation, we argue that the former produced a better trade-off between short-term cortisol secretion and long-term regulation: a more plausible strategy towards allostatic regulation than chronically elevated cortisol levels.

*D. Limitations and Future Work*

Our current model does not account for the maladaptive consequences of prolonged stress (cortisol), or the metabolic cost associated it. Work done by [31] has indeed found that accounting for these effects can have significant effects on an agents' action selection and homeostatic regulation. In future work, we will extend this current model by accounting for similar physiological (metabolic) effects of cortisol. We also do not claim that our relatively simple models of cortisol secretion and its allostatic functions are biologically accurate in the current model. This modelling currently serves as abstractions of their biological function. We aim to further extend the generative model beyond the simple implementation here and extend this computational model to physically embodied agents (i.e. robots) situated in a real, dynamic contexts. We believe that the simulated work undertaken here is an important first step in establishing the feasibility of such models in computational environments.

## VI. CONCLUSION

Both allostasis and the active inference framework seek to explain long-term regulation of biological systems in terms of the minimization of prediction errors through anticipatory adjustments of physiology and action. Their divergence in explanation lies in the underlying means of adapting to prediction errors: the latter emphasizes updates of statistical contingencies of a generative model, whereas the former emphasizes lower-level, hormonally induced (affective) adjustments to physiology. In the context of developing biologically-inspired models of adaptive, affective regulation for artificial systems, we propose a computational model that couples measures related to prediction errors from AIF to affective, stress-related hormonal dynamics (cortisol) proposed by allostasis. We evaluate this model with a simulated active inference agent endowed with an artificial physiology and incorporate two different allostatic functions of stress (cortisol). We found that these adaptive mechanisms of affective stress (via cortisol) provide benefits to the regulation of an artificial physiology of an active inference agent in a stochastic environment. We therefore argue that the coupling of information-theoretic prediction errors to low-level, biological hormonal dynamics associated with stress can provide a computationally efficient model to long-term regulation for embodied intelligent systems.

## ETHICAL STATEMENT

All simulations were run on a Windows 11 laptop, using an AMD Ryzen 7 6800 (3.2Ghz, 16-core) CPU, and 16GB DDR5 RAM. We estimate a total running time of simulations to be 90 minutes of total runtime (2-3 seconds per simulation), including the final set of simulations included in this paper, formulation of the initial generative model, and other pre-experiment validation and exploration of parameters. We calculate the CO2e impact of running our simulations using www.green-algorithms.org. We report the output as follows: this algorithm runs in 1h and 30min on 16 CPUs Ryzen 7 6800, and draws

**491.88 Wh**. Based in the United Kingdom, this has a carbon footprint of **113.68 g CO2e**, which is equivalent to **0.12 tree-months** (calculated using green-algorithms.org v2.2). We encourage anyone intending to replicate or extend on this work with additional simulations to consider energy-efficient ways to run them, e.g. by using the Climate-Aware Task Scheduler (https://github.com/GreenScheduler/cats) or similar tool. No human participants were used in this study. No datasets were used in this study. No ethical approval was required for this study. Our simulation results are publicly available in our supplementary material (we remove the link to our GitHub for the double-blind review process). The code used in this model will be made publicly available in the final version of this paper.